# ACES: A Teleoperated Robotic Solution to Pipe Inspection from the Inside

Eric Lucet* and Farès Kfoury
Université Paris-Saclay, CEA List, F-91120 Palaiseau, France
{eric.lucet, fares.kfoury}@cea.fr

**ABSTRACT**

This paper presents the definition of a teleoperated robotic system for non-destructive corrosion inspection of Steel Cylinder Concrete Pipes (SCCP) from the inside. A general description of in-pipe environment and a state of the art of in-pipe navigation solutions are exposed, with a zoom on the characteristics of the SCCP case of interest (pipe dimensions, curves, slopes, humidity, payload, etc.). Then, two specific steel corrosion measurement techniques are described. In order to operate them, several possible architectures of inspection system (mobile platform combined with a robotic inspection manipulator) are presented, depending if the mobile platform is self-centred or not and regarding the robotic manipulator type, namely a basic cylindrical manipulator, a self-centred one, or a force-controlled 6 degrees of freedom (DoF) robotic arm. A suitable mechanical architecture is then selected according to SCCP inspection needs. This includes relevant interfaces between the robot, the corrosion measurement Non-Destructive Testing (NDT) device and the pipe. Finally, possible future adaptation of the chosen solution are exposed.

**Keywords:** Robotic NDE; steel cylinder concrete pipes; remote inspection.

## 1. INTRODUCTION

This work is carried out in the scope of the ACES European research project. The mission of ACES is to assess the integrity of concrete infrastructure of nuclear power plants. In particular, one task focuses on inspection of SCCP from the inside by means of robotic mobile manipulator, aiming to assess chloride induced corrosion of steel in SCCP. Several types of inspection are usually carried out from the outside whenever possible, as they are confined areas, difficult to access by an operator. However, since steel corrosion initiates on internal surface of the steel liner, it is necessary to perform inspection from the inside of pipes to detect corrosion on early stages, before waterproofing provided by the steel liner is broken. Within this work, a prototype of a mobile robotic platform suitable for carrying out steel corrosion inspection from the inside of SCCP is designed, implementing a NDT technique.

The definition of such a mobile manipulator depends on mainly two critera, namely:
1) The inspected section geometry and its changing, like T sections, elbows, diameter change, etc.
2) The device and the process of corrosion measurement with their constraints; for example the acquisition time and the footprint of one measurement, the number of measurements in one position, the need of contact, angle position, etc.

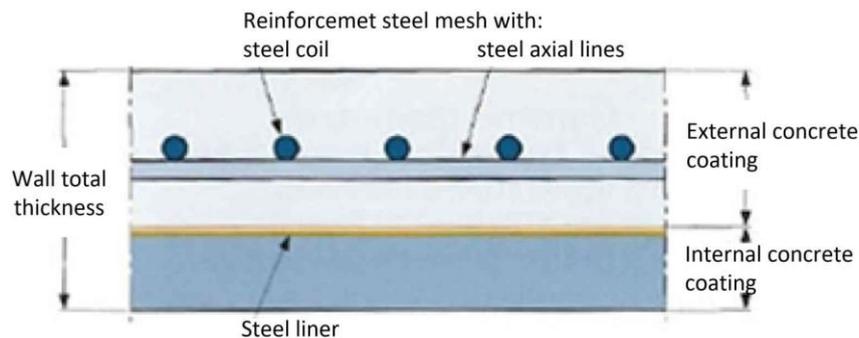
Figure 1. SCCP structure diagram

## 2. STATE OF THE ART IN-PIPE NAVIGATION SOLUTIONS

### 2.1. In-pipe environment

If many solutions for navigation in pipe exist, it is mainly because of the different possible geometries of the navigation environment and the necessary functionalities. Indeed, the network may vary in diameter range, material, fluid inside, with different joint configurations. Furthermore, each of these in-pipe configurations could be found in any orientation and possibly even consecutively, for example encountering two consecutive bends. Developing a single mobile platform to solve all of these problems in a wide range of diameters would require a fleet of multiple systems in different class sizes (example of modular and articulated-segmented design solution in [1]. Similar issues would be encountered for a robotic arm. Common encountered in-pipe bends and joints are the following [2]:

- Horizontal constant-diameter sections are the baseline for in-pipe complexity.
- Changes in diameter are a common occurrence; many robots take measures to prepare for this obstacle specifically.
- Elbows are commonly encountered and are often described in terms of their bend radius; lower radius bends being harder to navigate.
- T-Sections are challenging obstacles due to their lack of wall support; only sophisticated robotic platforms can navigate these (example of suitable two wheel chains mobile platform design in [3]).
- Vertical sections require a traction method that must also overcome gravity.
- Valves are particularly difficult because such plug valves can split the cross-section in two, obstructing full passage robots.

The context of this study includes the four first configurations, the last two being out of scope. It is in the light of these types of constraints and also of targeted applications that possible designs of mobile platforms and robotic arms are considered.

Preventive corrosion measurement is to be performed in areas where it is constrained to send people, to plan in advance the replacement of sections or repairing that will limit the deterioration of the concrete. Curative aspects are not considered here.

Considering an operational SCCP pipe network, different configurations might be encountered:

- High percentage of humidity (95 %) inside the pipes which, apart from maintenance operations, are usually filled with salt water.
- Slope up to 30° (vertical section parts not considered).
- Pipe inner diameter range from 600mm to 1000mm with a conical connecting part of a maximum angle to be set (example of this angle in blue Figure 2).
- 90° horizontal elbows with a minimum radius of curvature of about 3 times the pipe diameter (known as 3D Elbows).
- Passage by junctions such as T-sections or merging area (only navigation, no measurement).
- Hard bumps and irregularities up to 25mm thick at pipe joints.

Last constraint is especially true since the measurement at these joints is necessary. Start and end joints as well as the inside of bends and diameter change cones are also critical measurement areas. In addition, clearance issues are to be investigated with regard to these irregularities, which may also be caused by vegetation in pipes.

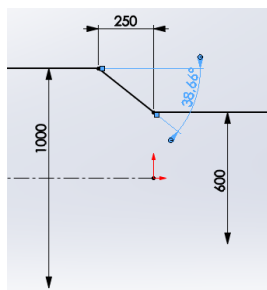

Figure 2. Lateral section of a diameter-varying pipe (lengths in mm)

## 2.2. Sate of the art of locomotion solutions

The pipe considered here is a SCCP as part of a nuclear power plant's concrete infrastructure. However, the need would remain similar in oil and gas pipelines, water pipelines, sewerage system or any piping systems that require specific inspection.

Main remaining weaknesses of existing solutions to be overcome are their maneuverability, meaning difficulties maneuvering inside a pipe with different diameters, curves or bends, and T joints, causing robots to be regularly found stuck during an operation because of a lack of stability and motion efficiency. Then, such issues of geometric changes in the pipe have also to be addressed.

We do not consider here specifically pipe climbing robots in vertical pipes, a review of 41 of such systems being available in [4]. Similarly, outer-pipe robots are not a topic of interest here, focusing on in-pipe robots. On the other hand, some robotic solutions designed for pipe diameters whose values are outside our scope are reported anyway, as relevant mechanical architecture examples can still be considered to be used at different diameters.

The first objective of the robotic platform is to be able to move inside a cylindrical pipe, meaning to generate traction in such an environment. For that, four in-pipe traction methods are possible [2]. Firstly, the gravity to propel on the floor restricts to only horizontal and slightly inclined pipes. Mass of the robot, as well as the geometry and composition of its contact zone affect the in-pipe wall adhesion. Secondly, wall-pressing consists of using the reaction force from the enclosed walls, usually in combination with a diameter adjustment mechanism. Third, wall magnetic adhesion is possible in ferrous pipelines that allow the production of reaction force through magnetics. Fourth, fluid Flow with the use of the transport medium to move, usually in combination with a passive Pipeline Inspection Gauge (PIG) or propeller device. In the context of this study, only the first two solutions are available, the pipes being made of concrete with an integrated steel sheet that is difficult to access for a reaction, and the measurements being made in a pipe emptied of any fluid.

Thus, based on a recent article [5], that is an extension of previous surveys on in-pipe robotic approaches for the inspection of unpiggable pipelines [2], on water in-pipe inspection robots [6], on in-pipe inspection robots from 1994 to 2010 [7], and on in-pipe hybrid locomotion robots from 1994 to 2012 [8], and completing it with other scientific papers about large-diameter in-pipe robots [9] and industrial solutions, a review of possible types of in-pipe inspection mobile robotic systems is proposed in the aim of direct to a choice of mobile platform solution for our needs. In addition, this gives inputs for a more complex optimized next generation inspection robotics system solution. This review is presented in the following by mobile platform architecture categories.

2.2.1. Wheel type robot

Wheeled robots are the simplest method of in-pipe locomotion. They are the most prevalent method because of their adaptability and ease of combination with other locomotion types, excluding tracks, to create hybrid in-pipe systems. Wheeled systems are also predominantly used with wall-press traction methods.
One of the first wall-pressing systems was the MOGRER in-pipe robot, developed by Niigata University around 1987 [10] for climbing angled pipes against gravity, from 520mm to 800mm in-pipe diameter. It was an improvement of the previous FERRET-1 design with the spring system forming a scissor structure similar to a pantograph mechanism (see Figure 3). Only one of the three wheels is actuated, and the robot is turning automatically following the pipe shape by wall-pressing. As another example of wheeled robot architecture, the two-modules NTU-Navigator robot was developed by National Taiwan University [11]. Its driving module consists of a worm gear, a DC motor and a driving wheel, and its steering module is composed of a servo motor and two steering wheels. The height of the robot is self-adjustable by a spring linked to the upper wheel (see Figure 4) for in-pipe diameter from 180mm to 250 mm. A concept from the University of California Irvine (see Figure 5) is proposed to apply a carbon-fiber coating to the in-pipes walls in order to fix them. A sensor is used to measure contact pressure against the pipe wall.

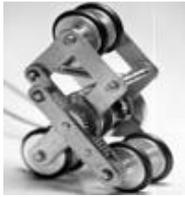
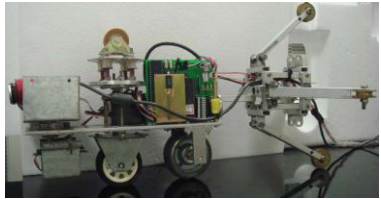
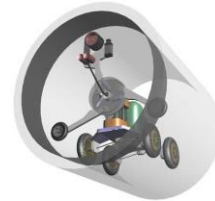

Figure 3. MOGRER in-pipe robot [10]  Figure 4. NTU-Navigator robot [11]  Figure 5. UCI pipe repair robot

One can also cite the MRINSPECT robots for 130mm to 180mm in-pipe diameter (see Figure 6) from Sungkyunkwan University of Suwon, Korea [12], or the flat two wheel chains robot [3] with one steering wheel and one driving wheel at each wheel chain, suitable for T-sections from 80mm to 100mm in-pipe diameter (see Figure 7).

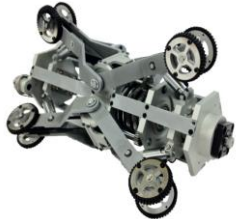
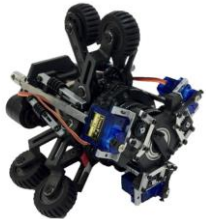
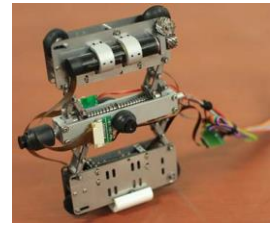

Figure 6. MRINSPECT VI and VI+ robots [13]  Figure 7. Flat robot with two wheel chains [3]

The SVM-RS robot [14] was designed by the Fraunhofer in Germany to clean and measure damage of large concrete pipes diameter between 1600mm to 2600 mm. It is four-wheel driven (see Figure 8). Its three degrees of freedom servo-controlled arm (see Figure 9) can pivot around a main axis and has two independent telescopic arms, a nozzle bank on the upper end of the arm and an ejector nozzle on the lower end. Navigation sensors are cameras and position relative to the pipe measuring sensors including inclinometer and seven light section sensors that use laser lines. The robot is linked to a cable with fibre optics and electrical wires for communication and energy, that is reinforced to be able to pull it in case of failure.

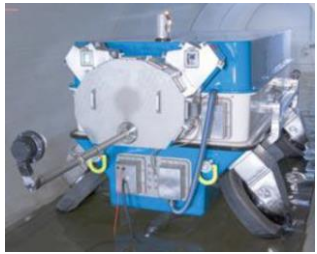
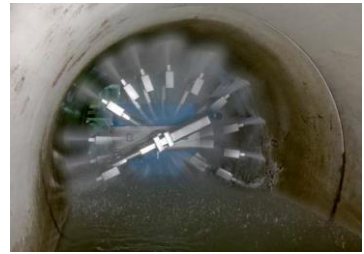

Figure 8. SVM-RS robot front view [14]  Figure 9. SVM-RS cleaning tool at rear [14]

2.2.2. Track / caterpillar / crawler type robot

Tracked robots, also known as caterpillars or crawlers, are used as an alternative to wheeled systems. Their large surface contact area generates high friction and reduce chances of losing wall contact. Tracks are more stable but also generally larger than wheels. They are used mainly for wall-pressing, and also for gravity alone. A classical design can be illustrated with a sewer cleaning and inspection robot (see Figure 10) designed by the department of Mechatronics of the University of Technical Education of Ho Chi Minh City, Viet Nam [15]. It is able to work underwater in pipes with diameters ranging from 300mm to 600 mm. As a more complex example, Paroys-II robot [16] is using an actively controlled pantograph mechanism with a partially passive spring mechanism, and a second set of articulated caterpillar tracks for an increased adaptive range from 400mm to 700mm in-pipe diameter.

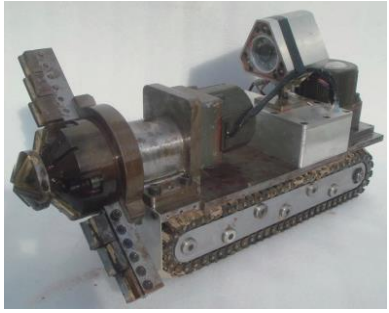
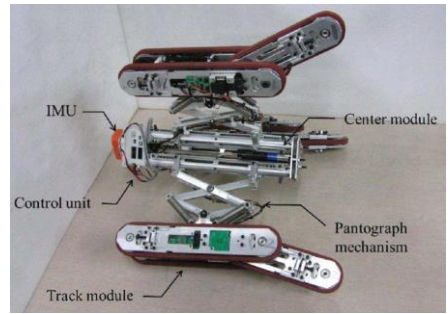

Figure 10. Sewer cleaning and inspection Robot [17]     Figure 11. Paroys-II robot [16]

A pipeline cleaning robot (see Figure 12 and Figure 13) from the Chinese Wuhan University of Technology [18] is able of self-adaption to variable cross-section pipe diameter from 150mm to 450mm by using an umbrella-like and lifting structure. Touch sensors at umbrella arms extremity are used for a closed-loop regulation of both lift and open-and-close motors. Also, the stability of the robot body posture is ensured by using a three-arms tail positioning empennage system. A strain gauge measuring the pressure is pasted on the robot tail to collect the tail curvature signal for controlling the tail open-closed angle.

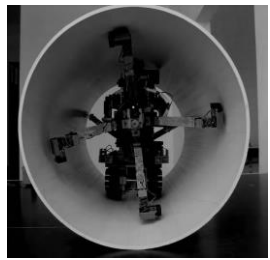
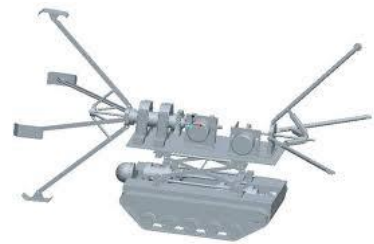

Figure 12. SVM-RS robot front view [18]     Figure 13. SVM-RS cleaning tool at rear [19]

The ILI crawler (see Figure 14 and Figure 15) from US Diakont company [20] is equipped with two Pulsed Eddy Current (PEC) probes in contact of the internal surface of a metallic pipe, with attached wheels and connected to a rigid, mechanical arm that extends from the crawler. The two arms rotate 180 degrees of the pipe circumference to cover the entire internal surface. Collected probe data, along with on-board camera images, operator's navigational instructions, and electrical power, are transmitted via an umbilical cable. It is able to support inspections for piping with diameters of 24 to 55 in (600mm to 1400 mm), being demonstrated on site in an operator's 36- and 42-in (900mm to 1100mm) diameter test cylinders.

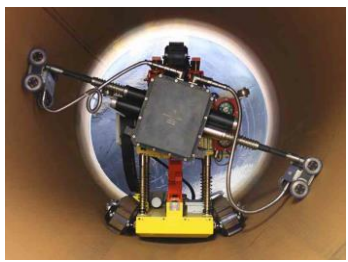
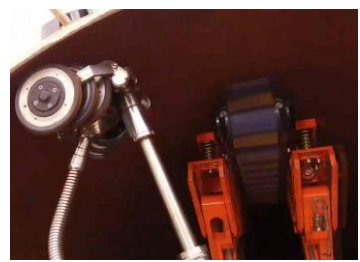

Figure 14. Diakont inline inspection (ILI) robotic crawler with 2 PEC probes inside a 42 in (1067 mm) diameter metallic pipe

Figure 15. Zoom on a probe and the top crawler deployed against the pipe inner surface

2.2.3. Screw / helical type robot

Screw robots use a rotary motion to move forward along a spiral inspection path in a pitched circle. They are always wall-pressing and generally difficult to move backward due to their angled wheels or tracks. They perform well in vertical sections and are resistant to slip due to their angled approach, even against an in-pipe flow. As an example, Shenyang University designed a robot capable of 250mm to 300mm pipeline diameter exploration using a passively adaptive four bar linkage [21] (see Figure 16). Or the China Nuclear Power Engineering Company of Shenzhen designed a screw-type drive unit [17] for 260mm in-pipe diameter, able to handle large payload (see Figure 17).

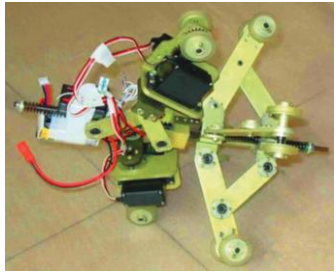 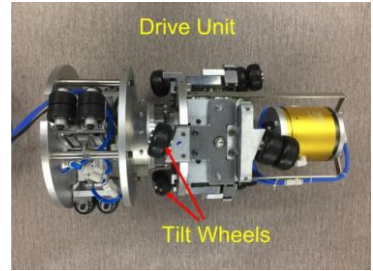

Figure 16. Helicoidal robot [21]     Figure 17. Screw-type drive robot [17]

### 2.2.4. Snake type robot

Snake robots take advantage of the length of the pipe. They are generally modular and adaptable to many in-pipe environments. They are composed of articulated modules equipped with wheels or tracks for locomotion. Such articulation allows many degrees of freedom making these robots very versatile in their approach to obstacles. For example, the BioInMech laboratory of the department of Robotics from Faculty of Science and Engineering of Ritsumeikan University, Japan, developed the AIRo-5.1 robot [22] consisting of two passive compliant joints and a single active compliant joint that is actuated by a series elastic actuator (see Figure 18), for 100mm to 130mm in-pipe diameter. Or the SINTEF ICT department of applied cybernetics of Trondheim, Norway, designed the PIKo snake-like robot [23] for horizontal and vertical pipes of varying diameter (see Figure 19).

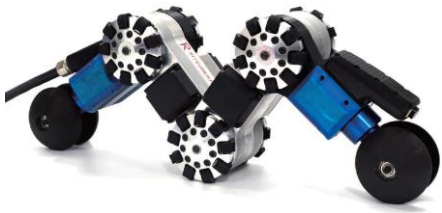 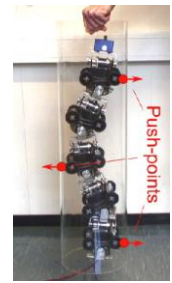

Figure 18. AIRo-5.1 snake robot [22]

Figure 19. PIKo snake-like robot [23]

### 2.2.5. Inchworm type robot

Inchworm robots are only wall-pressing, they generate traction through large normal force applied at a front or back module while a central module alternately contracts and extends. The use of point contact and the removal of wheels or tracks makes them more stable than other designs. They are slower than other types but can generally carry higher payloads due to their need for high wall-traction forces. They are useful in industrial transport tasks where speed is not important. The school of mechanical engineering from Yonsei University of Seoul, Korea, designed a steerable inchworm type in-pipe robot [24] dimensioned for 205mm to 305mm in-pipe diameter with the ability to traverse T-sections (see Figure 20). Another kind of design is the Compliant Modular Mesh Worm (CMMWorm) robot [25] designed by the Biologically Inspired Robotics laboratory of the Department of Mechanical Engineering of Case Western Reserve University, Cleveland, USA, that creates peristaltic motion with a continuously deformable exterior surface, dimensioned for 180mm to 220mm in-pipe diameter (see Figure 21).

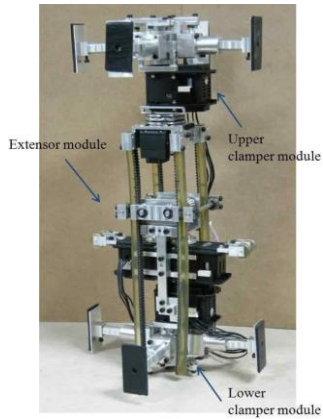
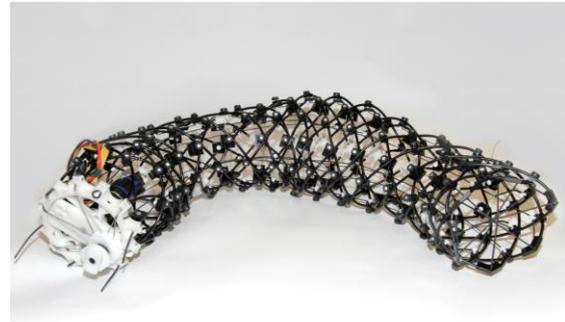

Figure 20. Inchworm inspection robot [24]

Figure 21. CMMWorm robot [25]

### 2.2.6. Walking type robot

Walking robots use multiple DoF legs to move inside the pipe. Their end effectors have low surface areas, which is useful for passing through in-pipe wall contaminants. Their wall-pressing functions sacrifice mobility for increased stability, being generally slower but with the ability to deliver heavier payloads. MORITZ robot, built at the Technical University of Munich, was one of the first walking style robots [26], capable of travelling through 600mm to 700mm diameter pipes and hold up to 20 kg payloads (see Figure 22). One can cite also more recent legged robots developed for more general robotics purposes, as for example the Mini Cheetah 3 from the MIT Biomimetic Robotics Laboratory that is a research prototype developed in 2018 [27].

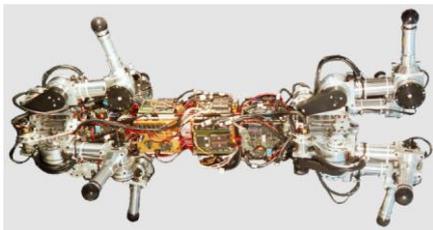
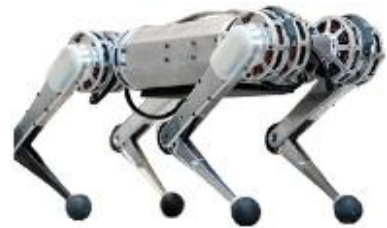

Figure 22. MORITZ pipe crawling robot [26]     Figure 23. Mini cheetah 3 robot from MIT [27]

### 2.2.7. Hybrid wall-pressing

Previous basic locomotion elements are often used in conjunction with another, forming a hybrid system, wall-pressing being the most popular method of generating in-pipe traction. These systems mostly consist of a chassis that is kept concentric with the pipe, using usually tracks or wheel locomotion subsystems that are mounted perpendicular to the chassis. Such a concentric system would be an advantage for the integration of a robotic manipulator that would no longer need to adapt its distance to the pipe inner wall during its 360-degree rotation. However, the drawback is that such system would not adapt to pipe change in diameter, T-section, and also elbow in case of rigid longitudinal body.

### 2.2.8. Summary of the different in-pipe robotic systems

Then, Table 1 summaries the advantages and disadvantages of the pipeline robotic systems presented above.

Table 1 Summary table of in-pipe robotic systems

| Type | Advantages | Drawbacks |
| --- | --- | --- |
| Wheels | • Common design<br>• Low maintenance costs | • Complex steering mechanism<br>• Poor traction performance on slippery terrains |

|  | • Lower cost than crawler | • High centering over uneven terrain |
|---|---|---|
| Crawler | • Handles more aggressive terrain | • Less energy efficiency<br>• Expensive |
| Helical | • Relatively less actuators needed<br>• Able to move forward and backward<br>freely in small diameter pipe | • Low speed<br>• Mainly suitable for precise motion in small pipe<br>• Not suitable for complex pipes with low payload |
| Snake | • No need of motor for wheels<br>• Move horizontally | • Needs more energy for operation |
| Inchworm | • Light weight<br>• Cheap<br>• Carry higher payloads | • Low speed<br>• Poor reliability |
| Walking | • Able to cope with different in-pipe environment | • Relatively complicated for practical application<br>• Instability during locomotion<br>• Generally slower |
| Hybrid wall-pressing | • Able to move forward and backward in small diameter pipe<br>• Adaptable to small pipes with different diameters | • Calibration at the center of pipe is needed<br>• Mainly suitable for precise motion in small pipe<br>• Not suitable for complex pipes with low payload |

The main function of the navigation platform is to provide a pose (position and orientation) of the frame linked to the sensor manipulator in relation to a frame linked to the SCCP. The latter must be defined each time the navigation platform encounters a new bend or joint. Figure 24 displays examples of navigation platforms and SCCP frames for cylindrical and elbow-shaped SCCP. Specification of the manipulator system depends on the ability of the navigation platform to position itself relative to the SCCP, which depends on its own locomotion architecture.

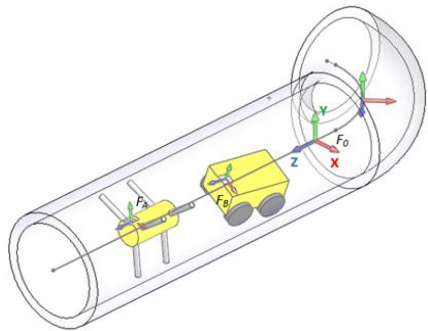

Figure 24. $F_A$ and $F_B$ frames provided by two navigation platforms

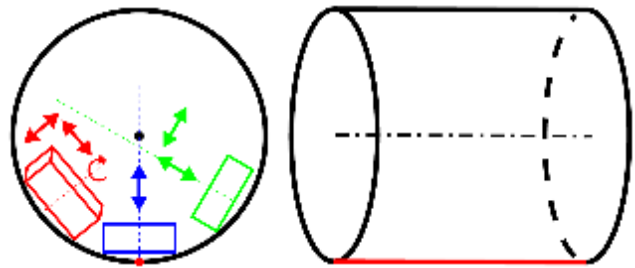

Figure 25. Robot position inside a pipe - cylinder front (left) and side (right) views

Thus, navigation platforms can be classified into two categories, whether they provide constant positions and orientation around lateral and vertical axes with respect to SCCP frame, meaning self-centred navigation platform (the central axis of the cylinder is the black point on the front view or the dotted line on the side view in Figure 25), or not, meaning non-centred navigation platform. The latter class of navigation platforms may be in different situations in the pipe, namely aligned with the pipe central axis and properly oriented (blue drawing on Figure 25), properly oriented (green drawing on Figure 25) or misoriented (red drawing on Figure 25).

## 3. CHOICE OF A ROBOTIC INSPECTION SYSTEM SOLUTION

### 3.1. Steel corrosion measurement process

In addition to environmental constraints, corrosion measurement NDT (see a review in [28]) is a key element for specifying the robotic system. This refers mainly to the measuring device and the complete measuring process description.

In particular, two steel corrosion measurement devices are considered. The first one is a rod electrode using potential field mapping technique (see an example Figure 6 in [29]). The second one is a commercial single-element Pulsed Eddy Current (PEC) probe. The selected measurement process, suitable for both techniques, is pointing and pressing with the measurement device over evenly spaced points along a 360° circle on the inner wall representing a cross section of the SCCP. The distance between measurement points is about 50mm. A cross section is said inspected when 3 circles are inspected, all located within a 200mm length of SCCP, that is about 150 inspection points per cross section. The next cross section to be inspected may be about 1 meter far from the previous cross section. Elbows being critical components (erosion, cracks and creep will most likely appear in elbows and at joints), it is necessary to perform some measurements there. Point by point process allows handling irregular areas or when a particular zone of interest needs to be inspected. Cross-sections do not have to be perfectly circular, and a bit elliptical inspection lines are acceptable. Contact of the sensor against the pipe inner wall is necessary, and inclination is to be avoided. A slight pressure on contact area should be assured by the robotic manipulator. These devices need a cable between them and the pipe entrance. Also, a screen outside is wired to the sensor for remote measurement activation and data observation.

### 3.2. Possible architectures for robotic inspection systems

Several possible architectures of inspection system (mobile platform combined with a robotic inspection arm) are presented, depending if the mobile platform is self-centered or not and regarding the robotic arm type, namely a basic cylindrical manipulator, a self-centered arm or a force-controlled 6 DoF robotic arm.

### 3.2.1. Cylindrical manipulator

This manipulator has a serial architecture, and comprises two actuated joints: one infinite revolution or at least with a range greater than one revolution, and a limited translation whose axis is perpendicular to the axis of the revolution joint. This architecture is the simplest and lightest for given sensor and SCCP specifications and may be typically associated with a navigation platform that centers the revolution joint axis in the SCCP, meaning a self-centered navigation platform. An example of such a cylindrical manipulator is given in Figure 9.

### 3.2.2. Self-centering manipulator

Compared to a cylindrical manipulator, the self-centring manipulator additionally includes an actuated centering system. The centering system is of the "wall-pressing" type with "legs" at the front and rear of the navigation platform pushing against the inner wall of the SCCP to center and orient the manipulator. For example, Figure 13 refers to such kind of system. The link between the centered manipulator and the navigation platform must remain passive, at least the 4 DoF around the lateral and vertical axes, or the whole system becomes hyperstatic in the SCCP. Then, this centering system is used as measurement manipulator only. Otherwise, with wheels it can become a self-centering navigation platform.

### 3.2.3. Force-controlled 6 DoF robotic arm

This manipulator is a serial robotic arm of at least 6 DoF including a contact sensor capable of detecting and locating contact points at its end effector with respect to its base frame. When the navigation platform carrying the manipulator stops at a point of interest, the manipulator detects 5 different contact points on the cylinder, provided that in any set of 4 points selected from those 5 points, the 4 points are not coplanar. Knowing the coordinates of these 5 points in the robot base frame, the inner cylinder of the SCCP is fully determined in this frame. Recognition of elbows or T sections may be more complex using this method. The measurement path, a circle whose center is on the axis of the inner cylinder, is then traveled by the measurement tool using the robotic arm. Despite its apparent genericity, drawbacks of such a solution are its high price and the complexity of control. Additional degrees of freedom may be required to control the arm internal configuration, for example to avoid collision situations with the navigation platform and/or the SCCP inner wall.

### 3.3. Chosen solution

The objective is a prototype mobile platform that allows the implementation of stable navigation functionalities in a pipe network, and carries a corrosion sensor manipulator. It must therefore have a sufficient payload, sufficient energy, one or more controllers with associated communication network, and a number of perceptual sensors of its environment. For all these reasons, and with a limited budget, it is out of scope to consider a commercial existing rover solution dedicated to the inspection of pipes in a nuclear environment, being waterproof and protected against radiation. Thus, according to SCCP inspection environment (see section 2.1) and process (see section 3.1), wheeled and tracked platforms may be suitable, the tracked solution allowing a better grip on the ground but also being more expensive and more energy consuming. Nevertheless, even if a complex solution, a legged mobile platform (see Figure 23) would still be an interesting category of solution to consider.

Then, in the particular case of navigation in a 600 mm to 1000 mm inner diameter pipe, a possible choice is the Clearpath Jackal [30] mobile robotic platform, this mobile base having a suitable size with sufficient clearance capacity (see Figure 26).

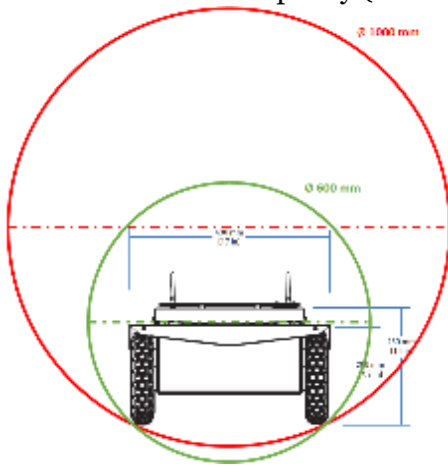
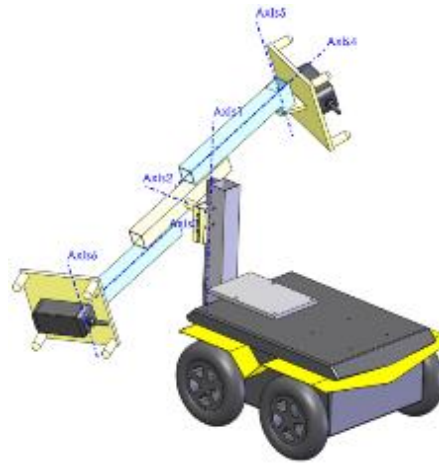

Figure 26. Jackal drawing in pipes of 600 mm and 1000 mm inner diameter

Figure 27. Manipulator integration into the mobile platform

For the manipulator specification, main challenges are appropriate measurement in elbow-shaped sections and in conical sections of pipes varying in diameter. With such requirements and if a wheeled navigation mobile platform is chosen, a simple cylindrical manipulator is not an option. Therefore, either a custom self-centering manipulator or a 6 DoF robotic arm is required.

If a commercial 6 DoF robotic arm is chosen, its generic mechanical architecture already physically allows the desired measurement areas to be reached, at least theoretically. Indeed, some situations of the end effector lead to unreachable joint positions due to contact between robot bodies and the SCCP inner wall. Thus, a redundant 7 or more DoF robotic arm may be necessary. On the other hand, the challenge is shifted to the perception and controller development, which need to robustly adapt to all types of configurations encountered.

A custom solution allows reduced complexity, both mechanically and for the control part, being specifically adapted to SCCP inspection requirements. Drawbacks of such solution are that complex pipe configurations would be difficult to manage and can also lead to a high number of degrees of freedom required, which will need to be controlled. As an example, the ILI robot manipulator architecture (see Figure 14) is not suitable for elbow pipes or conical parts.

Thus, finally the following customized mechanical architecture (see Figure 27) is proposed. For the correct centering of the manipulator, a vertical translation (Axis 1) is provided in relation to the mobile platform. This translation allows adaptation to a variation in pipe diameter. A lateral translation and a yaw rotation along the vertical axis that would also be needed for an accurate manipulator centering, are to be ensured by the mobile platform. Then, the Axis 2 horizontal rotation to be concentric with the pipe axis is actuated with an absolute rotation angle measurment. Axis 3 and 4 coupled radial translations are dedicated to the deployment of the 2 measuring end-effectors. They are compliant and reversible to adapt to surface roughness and allow sliding against the wall. Inclination of the two yellow plates, rotation movement about Axis 5 and Axis 6, perpendicular to the pipe axis, is symmetrical, and mechanically coupled, possibly operated by a motor. Any deviation from the tilt symmetry is subject to elastic return. Four contacts on ball joints are made between each plate and the pipe wall. This system, with 3 motors, can be centered in a tube, cone and elbow, and is compatible with pipe inner diameter in a range of 600mm to 1000 mm, allowing the use of 1 or 2 NDTs types simultaneously. The advantages of such a solution are its diameter adjustment capability and tolerance to surface irregularities, and its light weight. The drawback is that the rotation Axis 2 will not be merged with the axis of an elbow or conical pipe, the section covered by the manipulator being not fully orthogonal to the axis of a pipe with such a geometry. The magnitude of this orthogonality error will depend on the curvature of an elbow or the variation in diameter of a cone, with complementary point by point measurements still possible after a mobile platform shift.

## 4. CONCLUSIONS AND PERSPECTIVES

This paper deals with the design of a sensor manipulation robotic system dedicated to the measurement of corrosion from the inside of a pipe. Based on the specifications of the corrosion measuring NDT task to be performed and on a state of the art of existing systems, a robotic mobile manipulator solution is proposed as a compromise between reachability and adaptability on one side and overall weight and size, complexity and operability on the other side

This technical solution also allows future adaptation taking into account other likely scenarios of use. In particular, the current system performance may be improved by:
- 1 to 2 additional rotations to cope with elbows and conical parts in case of a correct roll, pitch, yaw positioning of the mobile platform;
- 2 additional rotations and 1 additional translation to cope with elbows and conical parts in case of roll and/or pitch and/or yaw positioning error from the mobile platform;
- Extra degrees of freedom to cope with high irregularities and holes.

## ACKNOWLEDGEMENTS

This work was carried out in the scope of ACES project. This project has received funding from the Euratom research and training programme 2014-2018 under grant agreement N° 900012.

## REFERENCES

[1] H. Schempf, E. Mutschler, A. Gavaert, G. Skoptsov and W. Crowley, "Visual and nondestructive evaluation inspection of live gas mains using the Explorer family of pipe robots," *Journal of Field Robotics,* vol. 27, p. 217–249, 2010.

[2] G. H. Mills, A. E. Jackson and R. C. Richardson, "Advances in the Inspection of Unpiggable Pipelines," *Robotics,* vol. 6, p. 36, December 2017.

[3] Y. Kwon, B. Lee, I. Whang, W. Kim and B. Yi, "A flat pipeline inspection robot with two wheel chains," in *2011 IEEE International Conference on Robotics and Automation*, 2011.

[4] P. Chattopadhyay, S. Ghoshal, A. Majumder and H. Dikshit, "Locomotion methods of pipe climbing robots: a review," *J. Eng. Sci. Technol. Rev. 11 (4),* p. 154–165, 2018.

[5] M. Z. A. Rashid, M. F. M. Yakub, S. A. Z. bin Shaikh Salim, N. Mamat, S. M. S. M. Putra and S. A. Roslan, "Modeling of the in-pipe inspection robot: A comprehensive review," *Ocean Engineering,* vol. 203, p. 107206, 2020.

[6]     J. M. Mirats Tur and W. Garthwaite, "Robotic devices for water main in-pipe inspection: A survey," *Journal of Field Robotics,* vol. 27, pp. 491-508, 2010.

[7]     I. N. Ismail, A. Anuar, K. S. M. Sahari, M. Z. Baharuddin, M. Fairuz, A. Jalal and J. M. Saad, "Development of in-pipe inspection robot: A review," in *2012 IEEE Conference on Sustainable Utilization and Development in Engineering and Technology (STUDENT)*, 2012.

[8]     N. S. Roslin, A. Anuar, M. F. A. Jalal and K. S. M. Sahari, "A Review: Hybrid Locomotion of In-pipe Inspection Robot," *Procedia Engineering,* vol. 41, p. 1456 – 1462, 2012.

[9]     W.-C. Law, I.-M. Chen, S.-H. Yeo, G.-L. Seet and K.-H. Low, "A Study of In-pipe Robots for Maintenance of Large-Diameter Sewerage Tunnel," *Proceedings of the 14th IFToMM World Congress,* January 2015.

[10]    T. Okada and T. Sanemori, "MOGRER: A vehicle study and realization for in-pipe inspection tasks," *IEEE Journal on Robotics and Automation,* vol. 3, p. 573–582, December 1987.

[11]    C. Lu, H. Huang, J. Yan and T. Cheng, "Development of a Pipe Inspection Robot," in *IECON 2007 - 33rd Annual Conference of the IEEE Industrial Electronics Society*, 2007.

[12]    S. U. Yang, H. M. Kim, J. S. Suh, Y. S. Choi, H. M. Mun, C. M. Park, H. Moon and H. R. Choi, "Novel robot mechanism capable of 3D differential driving inside pipelines," in *2014 IEEE/RSJ International Conference on Intelligent Robots and Systems*, 2014.

[13]    K. Sungkyunkwan University, *MRINSPECT robots*.

[14]    J. Saenz, N. Elkmann, T. Stuerze, S. Kutzner and H. Althoff, "Robotic systems for cleaning and inspection of large concrete pipes," in *2010 1st International Conference on Applied Robotics for the Power Industry*, 2010.

[15]    N. Truong-Thinh, N. Ngoc-Phuong and T. Phuoc-Tho, "A study of pipe-cleaning and inspection robot," in *2011 IEEE International Conference on Robotics and Biomimetics*, 2011.

[16]    J. W. Park, W. Jeon, Y. K. Kang, H. S. Yang and H. Park, "Instantaneous kinematic analysis for a crawler type in-pipe robot," in *2011 IEEE International Conference on Mechatronics*, 2011.

[17]    D. Zhu, H. Chen, D. Wang, W. Shu, D. Meng, X. Hu and S. Fang, "Design and Analysis of Drive Mechanism of Piping Robot," *Journal of Robotics and Automation,* vol. 3, November 2019.

[18]    Z. Li, J. Zhu, C. He and W. Wang, "A new pipe cleaning and inspection robot with active pipe-diameter adaptability based on ATmega64," in *2009 9th International Conference on Electronic Measurement Instruments*, 2009.

[19]    Z. X. Li, H. W. Li and Z. H. Li, "A New Air-Conditioning Pipeline Cleaning Robot System," in *Functional Manufacturing Technologies and Ceeusro II*, 2011.

[20]    K. R. Larsen, *Inspection Technology Assesses Unpiggable Pipelines,* 2020.

[21]    C. Ye, L. Liu, X. Xu and J. Chen, "Development of an in-pipe robot with two steerable driving wheels," in *2015 IEEE International Conference on Mechatronics and Automation (ICMA)*, 2015.

[22]    A. Kakogawa and S. Ma, "A Multi-Link In-Pipe Inspection Robot Composed of Active and Passive Compliant Joints," in *IEEE/RSJ International Conference on Intelligent Robots and Systems (IROS)*, Las Vegas, NV, USA (Virtual), 2020.

[23]     S. A. Fjerdingen, P. Liljeback and A. A. Transeth, "A snake-like robot for internal inspection of complex pipe structures (PIKo)," in *2009 IEEE/RSJ International Conference on Intelligent Robots and Systems*, 2009.

[24]     W. Jeon, J. Park, I. Kim, Y. Kang and H. Yang, "Development of high mobility in-pipe inspection robot," in *2011 IEEE/SICE International Symposium on System Integration (SII)*, 2011.

[25]     A. S. Boxerbaum, K. M. Shaw, H. J. Chiel and R. D. Quinn, "Continuous wave peristaltic motion in a robot," *The International Journal of Robotics Research,* vol. 31, p. 302–318, 2012.

[26]     A. Zagler and F. Pfeiffer, ""MORITZ" a pipe crawler for tube junctions," in *2003 IEEE International Conference on Robotics and Automation (Cat. No.03CH37422)*, 2003.

[27]     G. Bledt, M. J. Powell, B. Katz, J. Di Carlo, P. M. Wensing and S. Kim, "MIT Cheetah 3: Design and Control of a Robust, Dynamic Quadruped Robot," in *2018 IEEE/RSJ International Conference on Intelligent Robots and Systems (IROS)*, 2018.

[28]     R. F. Wright, P. Lu, J. Devkota, F. Lu, M. Ziomek-Moroz and P. R. Ohodnicki, "Corrosion Sensors for Structural Health Monitoring of Oil and Natural Gas Infrastructure: A Review," *Sensors,* vol. 19, 2019.

[29]     D. Luo, Y. Li, J. Li, K.-S. Lim, N. A. Mohd Nazal and H. Ahmad, "A Recent Progress of Steel Bar Corrosion Diagnostic Techniques in RC Structures," *Sensors,* vol. 19, no. 34, 2019.

[30]     Clearpath Jackal, [Online]. Available: https://clearpathrobotics.com/jackal-small-unmanned-ground-vehicle/.